%% file: main.tex
\documentclass[runningheads]{llncs}

\usepackage{eccv}

\usepackage{eccvabbrv}
\usepackage{multirow}
\usepackage{amsmath}
\usepackage{enumitem}
\usepackage{amssymb}
\usepackage{mathtools}
\usepackage{pifont}
\usepackage{wrapfig}
\usepackage{graphicx}
\usepackage{booktabs}
\usepackage{amssymb}
\usepackage{pifont}

\usepackage[accsupp]{axessibility}  
\newcommand{\ourmodel}{\textsc{DIM-Listener}\xspace} 
\newcommand{\pretraining}{Dyadic Interaction Modeling\xspace}
\newcommand{\pretrainingshort}{DIM\xspace}

\newcommand{\xmark}{\ding{55}}%

\usepackage{hyperref}

\usepackage{orcidlink}

\begin{document}

\title{\pretrainingshort: \pretraining for Social Behavior Generation}

\titlerunning{\pretraining}


\author{Minh Tran$^*$ \orcidlink{0009-0004-2391-3563} \and
Di Chang$^*$ \orcidlink{0009-0002-0281-8896} \and
Maksim Siniukov\and
Mohammad Soleymani\orcidlink{0000-0002-5873-1434}}

\authorrunning{M.~Tran et al.}

\institute{University of Southern California, Institute for Creative Technologies \\ \url{https://boese0601.github.io/dim/} \\
\email{soleymani@ict.usc.edu}\\}

\maketitle
\def\thefootnote{*}\footnotetext{equal contribution}

\begin{abstract}
Human-human communication is like a delicate dance where listeners and speakers concurrently interact to maintain conversational dynamics. Hence, an effective model for generating listener nonverbal behaviors requires understanding the dyadic context and interaction. In this paper, we present an effective framework for creating 3D facial motions in dyadic interactions. Existing work consider a listener as a reactive agent with reflexive behaviors to the speaker's voice and facial motions. The heart of our framework is Dyadic Interaction Modeling (DIM), a pre-training approach that jointly models speakers' and listeners' motions through masking and contrastive learning to learn representations that capture the dyadic context. To enable the generation of non-deterministic behaviors, we encode both listener and speaker motions into discrete latent representations, through VQ-VAE. The pre-trained model is further fine-tuned for motion generation. Extensive experiments demonstrate the superiority of our framework in generating listener motions, establishing a new state-of-the-art according to the quantitative measures capturing the diversity and realism of generated motions. Qualitative results demonstrate the superior capabilities of the proposed approach in generating diverse and realistic expressions, eye blinks and head gestures. The code is available at \url{https://github.com/Boese0601/Dyadic-Interaction-Modeling}.
\end{abstract}

\input{sec/1_intro}
\input{sec/2_related}
\input{sec/3_method}
\input{sec/4_exps}

\input{sec/5_conclusion}

\section*{Acknowledgement} This work is supported by the National Science Foundation under Grant No. 2211550. The work was also sponsored by the Army Research Office and was accomplished under Cooperative Agreement Number W911NF-20-2-0053. The views and conclusions contained in this document are those of the authors and should not be interpreted as representing the official policies, either expressed or implied, of the Army Research Office or the U.S. Government. The U.S. Government is authorized to reproduce and distribute reprints for Government purposes notwithstanding any copyright notation herein.



\bibliographystyle{splncs04}
\bibliography{main}

\input{X_suppl}

\end{document}

%% file: sec/1_intro.tex
\section{Introduction}
\label{sec:intro}

\begin{figure*}
\begin{center}
\includegraphics[width=\linewidth]{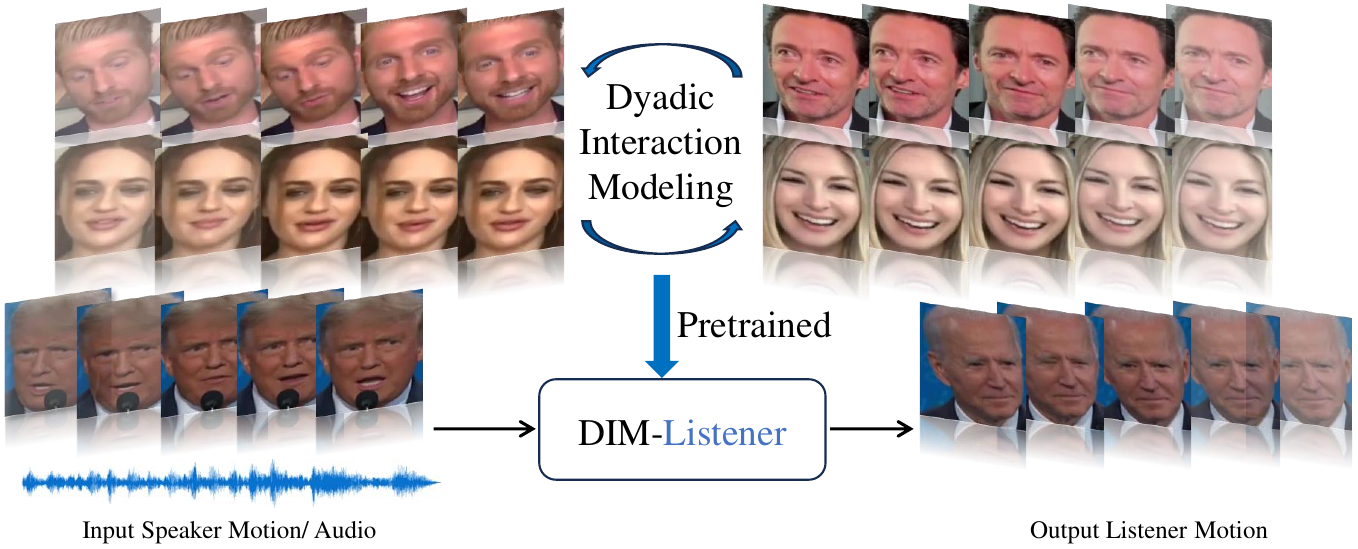}
\end{center}
  \caption{ We propose Dyadic Interaction Modeling, a pre-training strategy that jointly models speakers’ and listeners’ motions and learns representations that capture the dyadic context. We then utilize the pre-trained weights and feed multimodal inputs from the speaker into \ourmodel. \ourmodel is capable of generating photorealistic videos for the listener's motion.}
\label{fig:teaser}
\end{figure*}

Human nonverbal communication is like a dance. In human-human interaction, we concurrently encode and decode verbal and nonverbal messages, constantly attending to and reacting to our counterparts to maintain the flow of the conversation and engage in effective turn-taking \cite{devito2013interpersonal}. Hence, a socially intelligent agent needs to learn the interactive nature of nonverbal communication to create a seamless and compelling experience. In this paper, we present a method that goes beyond a reactive agent that only considers speakers' motions and voice by learning the listener-speaker dyadic context through self-supervised masked pre-training on a large dataset of human-human dyadic interactions, i.e., CANDOR \cite{reece2023candor}. 
As illustrated in Figure~\ref{fig:teaser}, We aim to generate a listener's lifelike and plausible head and facial movements in response to a speaker's motions and voice in a manner that resembles the contextualized bi-directional nature of human-human communication. 
Generating listener reactions has a wide range of applications, including in human-computer interaction~\cite{yu2017talking, liu2022audio,zhang2021flow}, virtual reality~\cite{jonell2020let,li2021ai}, the metaverse~\cite{chen2021high, cerekovic2016rapport, song2021fsft}, and media forensics~\cite{songadaptive, rossler2018faceforensics, song2022face, he2021forgerynet}. Given the ability of DIM to model dyading interactions, it is also able to produce realistic speaker motions.

Despite considerable progress, existing methods are limited, particularly in capturing real-life conversation dynamics, and difficult to generalize across listeners. 
Relevant work focused on either generating speakers' or listeners' movements. Speaker motion generation~\cite{chen2019hierarchical, song2021tacr, ji2021audio, eskimez2021speech, richard2021meshtalk, suwajanakorn2017synthesizing,chen2020talking, vougioukas2020realistic, song2021talking} primarily focuses on producing videos where lip movements are synchronized with speech, a.k.a., lip syncing. Listener motion generation, on the other hand, automatically interprets the speaker's voice and nonverbal behaviors to generate the listener's nonverbal responses. 
Traditionally, listener motion generation have not fully addressed the intricacies of dyadic interactions, which involve more than simple reactions to the speaker's utterances including a multifaceted, concurrent exchange of signals and emotions ~\cite{zhou2022responsive,ng2022learning,song2023emotional}. They were also sometimes tuned to specific listeners and unable to generalize to others \cite{ng2022learning}. Earlier work used heuristics and simple machine learning models to generate a limited set of reactions, e.g., head nods \cite{rapportAgent}. 
The Responsive Listening Head Generation (RLHG)\cite{zhou2022responsive} approach mimics the regression techniques found in Speaker Generation\cite{chen2019hierarchical}, leading to a reduction in the unpredictability of responses and a smoothing over of listener motions. 
Ng et al. ~\cite{ng2022learning} proposed Learning2Listen~(L2L) to mitigate this issue through a motion categorization strategy. In L2L, a codebook is learned through a VQ-VAE~\cite{mirsamadi2017automatic} that can be used to generate plausible and smooth behaviors. 
Emotional Listener Portrait (ELP)~\cite{song2023emotional} utilizes an Adaptive Space Encoder to map the discretized features combined with emotion to the listener's motion parameters. 
Similarly, Speaker motion generation techniques~\cite{richard2021meshtalk,fan2022faceformer,xing2023codetalker} have struggled to incorporate the feedback present in actual dyadic interactions. These shortcomings underscore a critical gap in the simulation of naturalistic human interactions, highlighting the need for a more holistic and dyadic modeling approach.

To address these challenges, we introduce \pretraining (\pretrainingshort), an innovative pretraining strategy designed to enhance a model's capacity to simultaneously encode a unified representation from both speaker and listener behaviors. This strategy draws inspiration from the recognition of the importance of bidirectional communication in interaction modeling, as demonstrated by Nojavanasghari et al.\cite{nojavanasghari2018interactive}, and Eskimez et al.\cite{eskimez2021speech}, who explored aspects of interactive emotion recognition and speech-driven facial animation, respectively. Moreover, the concept of integrating audiovisual information into a cohesive representation, as explored by Karras et al.\cite{karras2017audio} and Zhou et al.\cite{zhou2020makelttalk}, forms a foundational pillar for our approach, suggesting that a holistic and contextualized consideration of both audio and visual cues can significantly enhance animation quality.

Leveraging self-supervised contrastive learning and focusing on the reconstruction of masked-hidden units, \pretrainingshort captures the intricate exchanges characteristic of human dyadic conversations. This approach is inspired by the success of self-supervised learning techniques in capturing complex data patterns, as evidenced by the works of Chen et al.\cite{chen2020simple} and Gong et al.\cite{gong2022contrastive}, which have demonstrated significant advancements in data understanding without explicit labeling.

\pretrainingshort framework learns dyadic context for generating realistic motions. This framework enables the generation of detailed and lifelike facial expressions and head motions, marking a significant advancement in listener (\ourmodel) and speaker (DIM-Speaker) motion generation. Our method not only generates listener behaviors from speaker audio-visual inputs but could also adeptly produce speaker facial motions from speaker speech (DIM-Speaker), showcasing the framework's ability to model the bidirectional nature of human interactions.

Our main contributions can be summarized as follows:
\begin{itemize}
    \item We introduce \pretraining (\pretrainingshort), a pre-training strategy that enhances the model's ability to encode a unified representation from both speaker and listener behaviors through self-supervised contrastive learning, focusing on the reconstruction of masked-hidden units.
    \item Leveraging \pretrainingshort, we develop \ourmodel, a practical framework for listener motion generation in speaker-listener conversation with detailed and realistic facial expressions and head motions.
    \item \pretrainingshort is further leveraged to generate speaker facial behaviors from speaker speech (DIM-Speaker).
    \item Extensive experiments and visualizations on  motion generation demonstrate superior performance and effectiveness of our method.
\end{itemize}

%% file: sec/2_related.tex
\section{Related Work}\label{sec:related}

\subsection{Speech-driven Speaker Generation}

There is a wide body of work on automatic facial motion generation~\cite{weise2011realtime,li2013realtime,cao2016real,zollhofer2018state,kim2018deep,fried2019text,thies2020neural,lahiri2021lipsync3d,woo2023amii,kucherenko2023genea,song2023react2023,palmero2022chalearn,chang2023hierarchical}. A large number of studies focus on 2D facial animation~\cite{fan2015photo,chung2016out,chen2018lip,prajwal2020lip,vougioukas2020realistic,das2020speech,yi2020audio,chen2020talking,ji2021audio,zhou2021pose}. Traditional procedural techniques~\cite{massaro2012,taylor2012dynamic,xu2013practical,edwards2016jali} rely on a framework of explicit rules for simulating the movements of a talking mouth, such as the use of dominance functions~\cite{massaro2012} for delineating speech control parameters.


These procedural strategies offer precise control over mouth movement accuracy but are labor-intensive due to the extensive manual tuning required. In contrast, data-driven methods~\cite{cao2005expressive,liu2015video,karras2017audio,taylor2017deep,pham2018end,cudeiro2019capture,hussen2020modality,richard2021meshtalk,stan2023facediffuser} present an alternative, proposing various techniques to generate 3D facial animations with less manual intervention. Among these, Karras et al.~\cite{karras2017audio} propose an end-to-end network with linear predictive coding to generate audio and novel latent codes with facial expression variations. In Zhou et al.~\cite{zhou2018visemenet}, the proposed network predicts viseme curves by integrating phoneme groups, landmarks, and audio signals. VOCA~\cite{cudeiro2019capture} offers speaker-independent animation capturing diverse speaking styles, though it primarily affects the lower face. More recently, MeshTalk~\cite{richard2021meshtalk} has made strides in disentangling audio-correlated and uncorrelated facial motions through a categorical latent space. Innovations continue with FaceFormer~\cite{fan2022faceformer} and CodeTalker~\cite{xing2023codetalker}, which leverage long-term audio context via transformer-based models and augment them with a VQ-VAE motion prior for autoregressive motion synthesis, respectively.
\subsection{Speaker-driven Listener Generation}
Compared to Speaker Generation, Listener Generation pays more attention to the feedback of the listener's motion to the speaker. Ahuja et al. ~\cite{ahuja2019react} focus on the non-verbal body behaviors generation, ~\cite{bohus2010facilitating} and Greenwood et al.~\cite{greenwood2017predicting} study synchronized conversational agent motion in dyadic conversation adapt speech. Recent methods focus on generating 3D facial motions with additional inputs from the listener, such as text~\cite{chu2018face} or speech~\cite{jonell2019learning,jonell2020let}. Song et al.~\cite{song2023emotional} proposed ELP which learns the non-verbal listener motion in a dynamic communication. RealTalk~\cite{geng2023affective} retrieves possible videos of the listener's face with a large language model. The most similar work to our approach is that of Learning2Listen~\cite{ng2022learning}, which regresses the discrete listener head motion with VQ-VAE~\cite{van2017neural}. However, L2L model is unidirectional and does not consider the dyadic context. Unlike this work, the L2L \cite{ng2022learning} fixes the pre-trained VQ-Decoder to decode continuous signals from discrete motion predictions, we integrate the VQ-Decoder into motion prediction and optimize both discrete units (discrete latent codes) and continuous motions.

%% file: sec/3_method.tex
\section{Method}
\label{sec:method}
This section provides an overview of the proposed framework \ourmodel. We first delve into the facial motion representation EMOCA~\cite{EMOCA:CVPR:2021}, as discussed in Section~\ref{emoca}. Subsequently, we revisit VQ-VAE~\cite{van2017neural}, one of the core components in \ourmodel, in Section~\ref{vq}. We further detail our novel pre-training strategy, \pretraining, designed to foster the model's proficiency in developing a unified representation for both speakers and listeners in dyadic conversations. This is followed by an explanation of the fine-tuning stage for \ourmodel, presented in Section~\ref{dim}. We finally describe the rendering network~\cite{ren2021pirenderer}, transforming the estimated motions into realistic videos in Section.~\ref{render}.
\subsection{Facial Motion Representation}\label{emoca}
Following~\cite{ng2023text2listen}, we utilize EMOCA~\cite{EMOCA:CVPR:2021}, a 3D morphable model (3DMM), parameters as the representation for facial motions. EMOCA disentangles face shape, expression and head pose in three sets of parameters, allowing for capturing and expressing person-independent behaviors. Compared to DECA~\cite{DECA:Siggraph2021} used in previous work~\cite{learning2listen}, EMOCA can better capture facial expressions and head poses. Following prior work~\cite{ng2023text2listen,song2023emotional}, we extract camera, shape, expression and pose codes from the speakers and listeners frame by frame. Note that only expression code and pose code are used for representing human facial motions while camera code and shape code are only used for face mesh visualization purposes.  

\subsection{Learning discrete Speaker-Listener motions}\label{vq}
\begin{figure*}[t]
\begin{center}
\includegraphics[width=\linewidth]{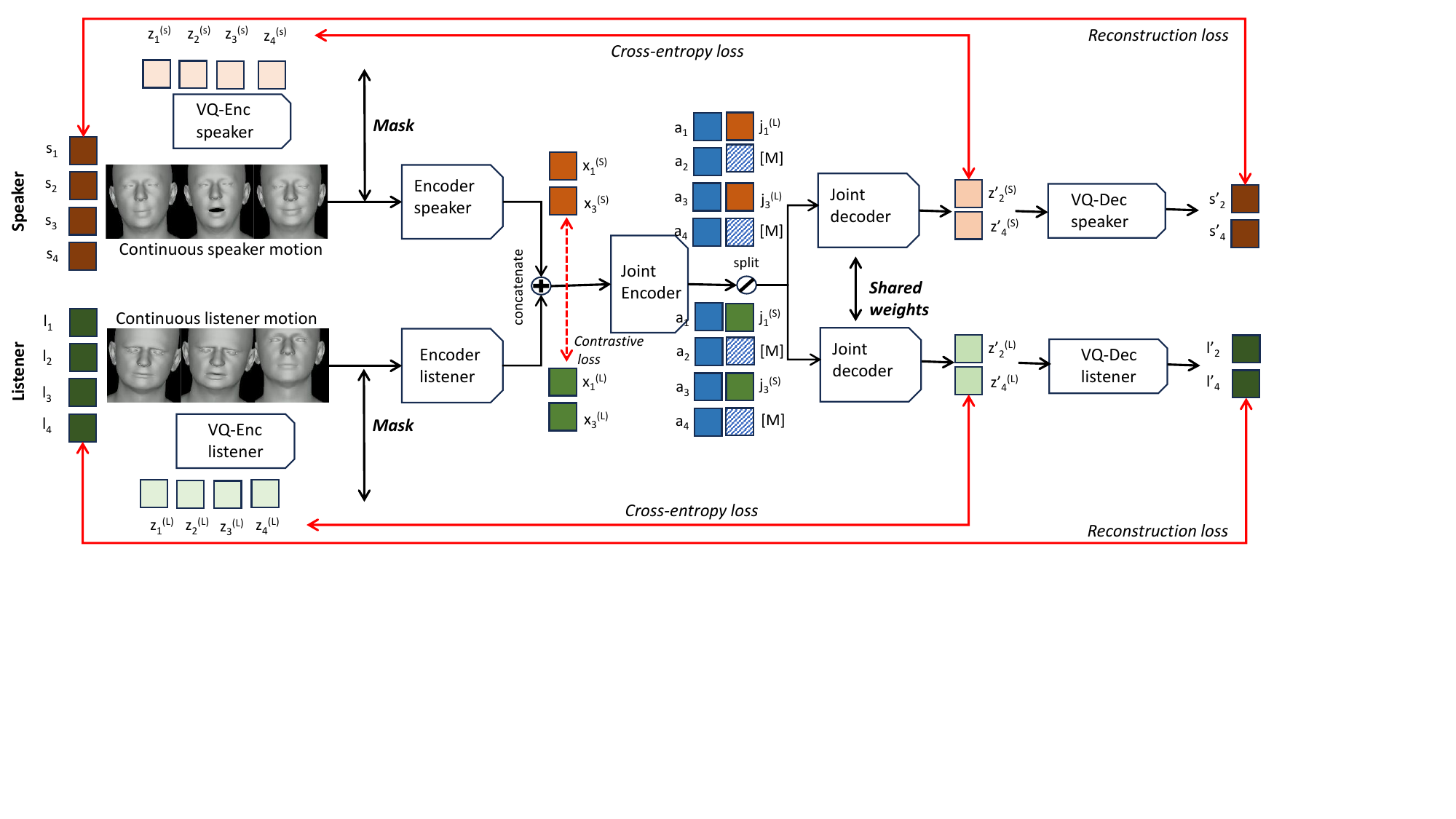}
\end{center}
  \caption{\textbf{\pretraining} learns a unified speaker-listener representation from dyadic interactions. 1) The framework takes both the ground-truth speaker motion  $s$ and the listener motion $l$ as input. 2) VQ-Encoders of speaker and listener then encode the motions to discrete units (discrete latent codes) $z^{(s)}$ and $z^{(l)}$. 3) The masked speaker's and listener's motions are further encoded and concatenated so that a unified representation is learned with contrastive loss. 4) Then, the split unified representation and speaker audio feature $a$  are decoded into discrete unit predictions $z'^{(s)}$ and $z'^{(l)}$ supervised by cross-entropy loss. 5) Finally, the generated speaker motions $s'$ and listener motions $l'$ are decoded from these discrete unit predictions to optimize the reconstruction loss.}
\label{fig:pipeline}
\end{figure*}

Formally, we train two separate VQ-VAE models, denoted as $VQ^{(s)}$ and $VQ^{(l)}$, to handle speaker and listener motions, respectively. For clarity, we present only the broad overview of the VQ-VAE workflow for encoding and reconstructing listener motions. In particular, the encoder $Enc^{(l)}_{VQ}$ processes the listener motions $l_{1: T}$ to produce the continuous latent representation $z^{(l)}_{cont}=Enc^{(l)}_{VQ}(l_{1: T})$. The latent representation is then quantized to $z^{(l)}$ using the nearest vector from the shared codebook $C$. The reconstruction of the listener motion is then given by:
$\hat{l}_{1: T} = Dec^{(l)}_{VQ}(z^{(l)}).$

$VQ^{(l)}$'s training objective incorporates both reconstruction and codebook losses. The reconstruction loss $\mathcal{L}_{recon}$ aims to minimize the difference between the original and reconstructed motions:
\begin{equation}
    \mathcal{L}_{recon}^{(l)} = \| l_{1: T} - \hat{l}_{1: T} \|_2^2,
\end{equation}
where $\| \cdot \|_2$ denotes the squared Euclidean norm. The codebook loss $\mathcal{L}_{codebook}$, comprising the commitment loss and the codebook utilization loss, is defined as:
\begin{equation}
    \mathcal{L}_{codebook}^{(l)} = \| sg[E(l_{1: T})] - z^{(l)} \|_2^2 + \beta \| sg[z^{(l)}] - E(l_{1: T}) \|_2^2,
\end{equation}
with $sg[\cdot]$ denoting the stop-gradient operator, and $\beta$ is a hyperparameter that balances the commitment of the encoded vectors to the codebook. The overall loss combines these components:
\begin{equation}
    \mathcal{L_{VQ}}^{(l)} = \mathcal{L}_{recon}^{(l)} + \gamma \mathcal{L}_{codebook}^{(l)},
\end{equation}
where $\gamma$ is a weighting factor for the codebook loss. 

Following L2L \cite{ng2022learning} and ELP~\cite{song2023emotional}, which leverage discrete motions to model the complex interplay in human interactions and animate listener motion from speaker audiovisual inputs, we use VQ-VAE~\cite{van2017neural} to learn discrete motion codebooks for \textbf{both speaker and listener} behaviors, and use the learned codebooks to generate discrete latent codes for our \pretraining process. 
\subsection{\pretraining}\label{dim}
\textbf{Notations.} We denote the input speaker motions as $s_{1: T}$, the input speaker audio features as $a_{1: T}$, the input listener motion as $l_{1: T}$, the encoders (and quantization modules) of the VQ-VAE models trained for speaker motions and listener motions as $Enc^{(s)}_{VQ}$ and $Enc^{(l)}_{VQ}$, respectively. We refer to the decoder of the VQ-VAE models trained for speaker motions and listener motions as $Dec^{(s)}_{VQ}$ and $Dec^{(l)}_{VQ}$, respectively. 

The overall pipeline is illustrated in Figure~\ref{fig:pipeline}. Given $s_{1: T}$ and $l_{1: T}$, \ourmodel first generates $z^{(s)}$ and $z^{(l)}$ as the discrete motion sequences produced by the learned motion codebooks: $z^{(s)}=Enc^{(s)}_{VQ}(s_{1: T})$ and $z^{(l)}=Enc^{(l)}_{VQ}(l_{1: T})$, and uses these sequences as the discrete latent codes to later train the network. It is important to note that both $Enc^{(s)}_{VQ}$ and $Enc^{(l)}_{VQ}$ are frozen to produce consistent discrete latent codes.
Then, we independently mask $p\%$ of random frames for both speakers and listeners, i.e.,
\begin{equation}
\begin{aligned}
    s_{masked}=Mask_p(s+E_s+E^P_s) \; ; \; l_{masked}=Mask_p(l+E_l+E^P_l)
\end{aligned}
\end{equation}
where $E_s$ and $E_l$ denote the learnable speaker and listener embeddings and $E^P_s$ and $E^P_l$ denote the sinusoidal positional embeddings. $s$ and $l$ are speaker motions and the listener motions from ground truth. \\
We then feed $s_{masked}$ and $l_{masked}$ to independent role-specific Transformer encoders $Enc^{(s)}$ and $Enc^{(l)}$ to produce $x^{(s)}_{masked}$ and $x^{(l)}_{masked}$. To further enhance $Enc^{(s)}$ and $Enc^{(l)}$ to extract robust features for speaker and listener inputs, we propose using contrastive learning to match speaker-listener pairs. We use the contrastive loss for the task
\begin{equation}
\begin{aligned}
    \mathcal{L}_c = -\frac{1}{N} \Sigma_{i=1}^N log [\frac{exp(x^{(s)}_i \cdot x^{(l)}_i /\tau)}{\Sigma_{i \neq k \;\;}exp(x^{(s)}_i \cdot x^{(l)}_k/\tau)+exp(x^{(s)}_i \cdot x^{(l)}_i/\tau)}]
\end{aligned}
\end{equation}
where $x^{(s)}_i$ and $x^{(l)}_i$ are the temporally mean-pooled feature representations of the $i^{th}$ sample of $x^{(s)}_{masked}$ and $x^{(l)}_{masked}$, respectively.\\
We further concatenated  $x^{(s)}_{masked}$ and $x^{(l)}_{masked}$ along the feature dimension and forward the fused feature into a joint Transformer encoder to produce $j^{(s-l)}_{masked}$ and later split $j^{(s-l)}_{masked}$ back into separate speaker and listener features, i.e.,
\begin{equation}
\begin{aligned}
    [j^{(s)}_{masked}\; ; \; j^{(l)}_{masked}]=j^{(s-l)}_{masked}=Enc^{(joint)}([x^{(s)}_{masked} \; ; \; x^{(l)}_{masked}])
\end{aligned}
\end{equation}
For the masked frames reconstruction task, we pad $j^{(s)}_{masked}$ and $j^{(l)}_{masked}$ with trainable mask tokens at the masked positions as $j^{(s)}$ and $j^{(l)}$. We further add learnable speaker and listener embeddings, along with sinusoidal positional embeddings, to $j^{(s)}$ and $j^{(l)}$ before concatenating them with extracted audio features for predicting the masked discrete motions via a joint Decoder.
\begin{equation}
\begin{aligned}
    z'^{(s)}=Dec^{(joint)}([j^{(l)}+E^D_l+E^p_l \; ; \; a_{1:T}])\\
    z'^{(l)}=Dec^{(joint)}([j^{(s)}+E^D_s+E^p_s \; ; \; a_{1:T}])
\end{aligned}
\end{equation}
The predicted discrete motions $z'^{(s)}$ and $z'^{(l)}$ are then forwarded into $Dec^{(s)}_{VQ}$ and $Dec^{(l)}_{VQ}$ for final continuous motion predictions.
\begin{equation}
\begin{aligned}
    \tilde{s}=Dec^{(s)}_{VQ}(z'^{(s)})\ \; ; \; \tilde{l}=Dec^{(l)}_{VQ}(z'^{(l)})
\end{aligned}
\end{equation}

For reconstruction, we train our network on both a cross-entropy loss on the discrete motions and a reconstruction loss
\begin{equation}
\begin{aligned}
    \mathcal{L}^{(s)}_{rec} = \Sigma_{t\in M} -log p (z^{(s)}_t | z'^{(s)}_t) + ||s_t-\tilde s_t||_2^2 \\
    \mathcal{L}^{(l)}_{rec} = \Sigma_{t\in M} -log p (z^{(l)}_t | z'^{(l)}_t) + ||l_t-\tilde l_t||_2^2
\end{aligned}
\end{equation}
where $M$ denotes the masked position during the masking process. Our final loss is defined as 
\begin{equation}
    \mathcal{L} = \lambda_1 \mathcal{L}_c + \lambda_2 (\mathcal{L}^{(s)}_{rec} + \mathcal{L}^{(l)}_{rec})
\end{equation}
where $\lambda_1$ and $\lambda_2$ are hyperparameters. 

\begin{figure*}[t]
\begin{center}
\includegraphics[width=\linewidth]{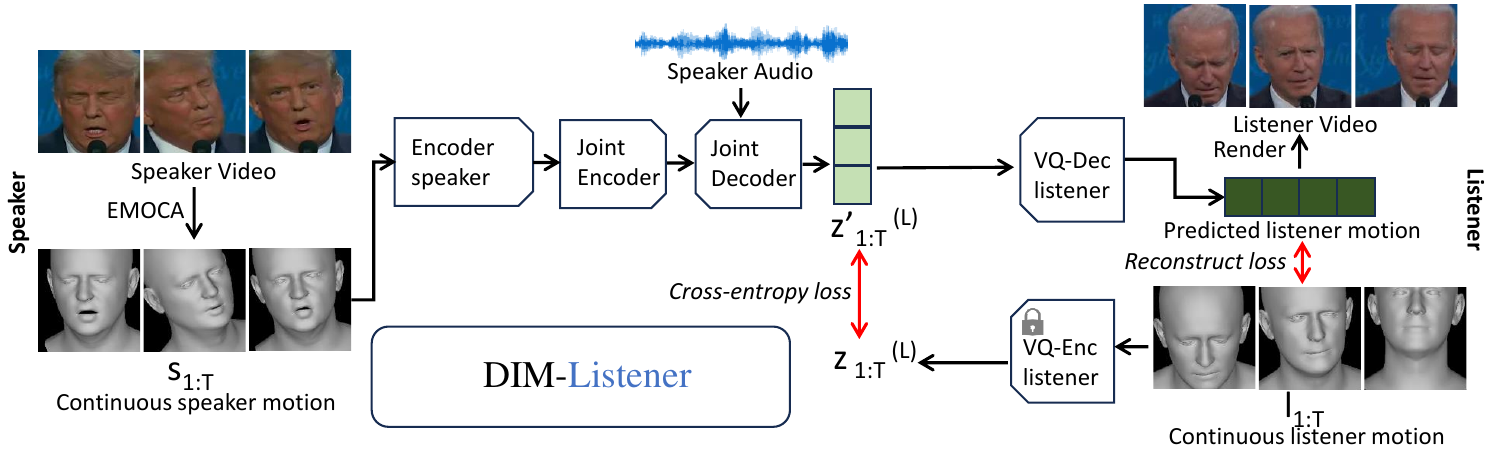}
\end{center}
  \caption{For fine-tuning the model on listener motion generation, speaker input is not masked, and the listener input is entirely masked. We train the framework with the same cross-entropy loss and reconstruction loss from \pretraining while keeping the weights of listener VQ-Encoder fixed.}
\label{fig:listener}
\end{figure*}
We then fine-tune the model for listener motion generation with initialized weights from \pretraining, as shown in Figure~\ref{fig:listener}. During the fine-tuning phase where $s_{1:T}$ and $a_{1:T}$ are available but $l_{1:T}$ is the target prediction, we do not mask $s_{1:T}$ (i.e., $p_s=0\%$) while mask entirely the listener input (i.e., $p_l=100\%$). The VQ-Encoders are frozen while the remaining weights are fine-tuned with the reconstruction loss of the listener motions $\mathcal{L}^{(l)}_{rec}$.

\subsection{Photorealistic Render}\label{render}
To visualize the real-human listener video from the predicted expression and pose EMOCA parameters, we re-train PIRenderer, a motion-to-video rendering network~\cite{ren2021pirenderer}. Collecting long-term person-specific portrait videos to train high-quality person-specific models is challenging and out of the scope of this work. Hence, we collect listener-speaker pairs and train a separate rendering network on pairs with the same speaker-listener identities. We feed the first frame of each video as the reference portrait image, together with the EMOCA parameters extracted from the later frames, into the renderer and train it with an image reconstruction loss. In this way, we use the predicted EMOCA motion prediction to drive the reference portrait and generate a vivid photorealistic video, as shown in the right side of Figure~\ref{fig:listener}.


%% file: sec/4_exps.tex
\section{Experiments}\label{sec:exp}
\subsection{Experimental Settings}

\noindent \textbf{Listener Behavior Generation.} We evaluate the speaker-driven listener motion generation of \ourmodel on  ViCo~\cite{zhou2022responsive} and LM\_Listener~\cite{ng2023text2listen} dataset.ViCo~\cite{zhou2022responsive} consists of 483 video sequences featuring 50 unique listeners. The dataset that was proposed in LM\_Listener~\cite{ng2023text2listen} is an extended version of the dataset introduced in Learning2Listen~\cite{ng2022learning} (L2L), which contains 2366 training segments, 222 validation segments, and 543 test segments of a single listener (Trevor Noah). \\
 \textbf{Speaker Behavior Generation.} We evaluate the speech-driven speaker motion generation on BiWi~\cite{fanelli2010}. BiWi~\cite{fanelli2010} comprises affective speech recordings and the associated high-resolution, dynamic 3D facial geometry data. It features recordings from 14 participants who were instructed to read 40 sentences in English, recorded at a frame rate of 25 frames per second. On average, each recorded sequence is approximately 4.67 seconds.\\
\textbf{\pretraining} is performed on CANDOR~\cite{reece2023candor} dataset. It consists of a collection of 1,656 conversations in English. With a total duration of over 850 hours, this large corpus includes more than 7 million words, encompassing rich audio, video, and speech transcripts. It enables detailed moment-to-moment analysis of vocal and facial expressions as well as semantic and conversational contexts. CANDOR provides speech diarization and alignment, which was used for speaker/listener designation at the utterance level. The dataset also provides separate speech tracks, so there is no overlapping speech.

\par \noindent \textbf{Implementation Details} For the pre-training stage, we use an Adam optimizer with a learning rate of $1e^{-5}$ and train the model on CANDOR \cite{reece2023candor} for $100$ epochs. We use a masking ratio $p=75\%$, which is determined with hyper-parameter tuning with respect to the pre-training validation reconstruction loss. We use HuBERT \cite{hsu2021hubert} as our audio feature extractor.
We use the same architectures for the Transformer encoders and decoders of \ourmodel and VQ-VAE, which consists of a transformer with $8$ hidden layers, $8$ attention heads and an intermediate size of $768$. For pre-processing of CANDOR, we use the provided transcript with the dataset to extract utterance-level samples of speaker-listener behaviors. For pre-processing of ViCo \cite{ECCV2022} and LM\_Listener \cite{ng2023text2listen}, we follow the pre-processing guidelines and train-test splits from prior works \cite{ng2022learning, ECCV2022, song2023emotional} for fair comparisons. 
For all experiments, we extract EMOCA~\cite{EMOCA:CVPR:2021}, an expressive 3DMMs parameter set, from visual data and adopt it as facial representation for all methods and re-train them for the sake of fair comparison. We follow previous work~\cite{zhou2022responsive,ng2022learning,song2023emotional} and adopt expression code (Exp) and pose code (Pose) of EMOCA~\cite{EMOCA:CVPR:2021} parameters as the description of the speaker and listener's motions. 
We use HuBERT \cite{hsu2021hubert} to extract audio features. Since HuBERT is already a robust pre-trained speech encoder, we do not use the audio features in our encoders and directly feed them into our model's decoder. For the implementation source code of our method, please see supplementary materials.

\par \noindent \textbf{Baselines} We compare the performance of \ourmodel on Listener Generation with random and nearest neighbor baselines in addition to the following state-of-the-art methods. 

\begin{itemize}
    \item RLHG~\cite{zhou2022responsive} is a regression method. We use their released official code for a fair comparison.
    \item L2L~\cite{ng2022learning} utilizes a discrete codebook to synthesize motion patterns. We use their released official code for a fair comparison.
    \item ELP~\cite{song2023emotional} extracts speaker-style features and emotion vectors. Due to the unavailability of the source code, we re-implement the baseline following the paper.  We provide our re-implementation in the supplementary materials.
\end{itemize}

The random and nearest neighbor baselines are also widely reported in previous work~\cite{zhou2022responsive,ng2022learning,song2023emotional}, which include:
\begin{itemize}
    \item We randomly select the facial and head motion parameters in the training data, and inject random small perturbations into the normal distribution.
    \item We use the smoothed speaker's motion as the listener's motion.
    \item  For arbitrary input speaker motion or audio input, we find its nearest neighbor from the training set and use its corresponding listener motion as output. 
\end{itemize}

\par \noindent \textbf{Metrics} We re-implement all baselines and train them with EMOCA 3DMM as output motion representation. We then evaluate them using the metrics reported in LM\_Listener~\cite{ng2023text2listen} and ELP~\cite{song2023emotional}, namely the  \textit{Frechet Distance} (FD), \textit{Paired FD} (P-FD) for synchrony , \textit{Mean Squared Error} (MSE), \textit{SI for Diversity} (SID), \textit{Variance} (Var), and \textit{Residual Pearson Correlation Coefficient} (rPCC). We provide more details about the used metrics in the supplementary materials.


\begin{table}[t!]
\centering

\caption{Quantitative Comparison of Listener Generation  on ViCo~\cite{zhou2022responsive} Dataset. $\dagger$ denotes the corresponding method has been pre-trained on the CANDOR\cite{reece2023candor} dataset. $*$ denotes the method didn't release any code, and we re-implement it from scratch on our own. }
\label{tab:quant_comp_vico}
\resizebox{\textwidth}{!}{
{\begin{tabular}{lcccccccccccc}

\toprule
\multirow{2}[2]{*}{Method}  & \multicolumn{2}{c}{\textbf{FD}$\downarrow$} & \multicolumn{2}{c}{\textbf{P-FD}$\downarrow$} & \multicolumn{2}{c}{\textbf{MSE}$\downarrow$} & \multicolumn{2}{c}{\textbf{SID}} & \multicolumn{2}{c}{\textbf{Var}}  & \multicolumn{2}{c}{\textbf{rPCC}$\downarrow$}  \\ 

\cmidrule(lr){2-3} \cmidrule(lr){4-5}  \cmidrule(lr){6-7}  \cmidrule(lr){8-9}  \cmidrule(lr){10-11}  \cmidrule(lr){12-13}
 &  \textbf{Exp}   & \textbf{Pose} & \textbf{Exp}   & \textbf{Pose}   & \textbf{Exp}   & \textbf{Pose}  & \textbf{Exp}   & \textbf{Pose} & \textbf{Exp}   & \textbf{Pose} & \textbf{Exp}   & \textbf{Pose} \\
\midrule
 Random & 72.88 &0.12 &  75.82 &0.12 & 2.05 & 0.03& 4.68 & 3.83 & 1.49& 0.02 & 0.09& 0.17\\
Nearest Audio & 65.77&0.10 &68.84 &0.10 &1.77 &0.03 &4.81 &3.93 & 1.49&0.02 & 0.07&0.11 \\
Nearest Motion & 42.41 &\textbf{0.06} &45.33 &\textbf{0.06} &1.27 &0.02 &4.62 &3.74 &1.49 &0.02 &0.08 &0.12\\
 ELP$*$~\cite{song2023emotional}&47.17 & 0.08&47.48 & 0.08&0.98& 0.02& 1.76&1.66& 1.49&0.02 & 0.09& \textbf{0.01} \\
 RLHG~\cite{zhou2022responsive} & 39.02&0.07 &40.18 &0.07  &0.86 &\textbf{0.01} &3.62 &3.17 &1.52 &0.02 &0.08 &0.02   \\

L2L~\cite{ng2022learning} & 33.93 & \textbf{0.06}&35.88 & \textbf{0.06}&0.93 &\textbf{0.01} &2.77 &2.66 &0.83 &0.02 &\textbf{0.06} &0.08 \\
L2L$\dagger$~\cite{ng2022learning} &31.03 &\textbf{0.06} & 33.02&\textbf{0.06} & 0.87&\textbf{0.01} &3.53 & 3.05&0.83 &0.02 &\textbf{0.06} &0.09  \\
GT & - & -& - & -& - & - &5.03 &4.07 &0.93 &0.01 &- & - \\
\midrule

\ourmodel$\dagger$  &\textbf{23.88} &\textbf{0.06} &\textbf{24.39} &\textbf{0.06} &\textbf{0.70} &\textbf{0.01} &3.71 &2.35 &1.53 &0.02 &\textbf{0.06} &0.03 \\
\bottomrule
\end{tabular}}}
\end{table}

\begin{table}[t!]
\centering

    \caption{Quantitative Comparison of Listener Generation on LM\_Listener~\cite{ng2023text2listen} Dataset. $\dagger$ denotes the corresponding method has been pre-trained on the CANDOR\cite{reece2023candor} dataset. $*$ denotes the method didn't release any code, and we re-implement it from scratch on our own.}
\label{tab:quant_comp_lm}
\resizebox{\textwidth}{!}{
{\begin{tabular}{lcccccccccccc}

\toprule
\multirow{2}[2]{*}{Method}  & \multicolumn{2}{c}{\textbf{FD}$\downarrow$} & \multicolumn{2}{c}{\textbf{P-FD}$\downarrow$} & \multicolumn{2}{c}{\textbf{MSE}$\downarrow$} & \multicolumn{2}{c}{\textbf{SID}} & \multicolumn{2}{c}{\textbf{Var}}  & \multicolumn{2}{c}{\textbf{rPCC}$\downarrow$}  \\ 
\cmidrule(lr){2-3} \cmidrule(lr){4-5}  \cmidrule(lr){6-7}  \cmidrule(lr){8-9}  \cmidrule(lr){10-11}  \cmidrule(lr){12-13}
 &  \textbf{Exp}   & \textbf{Pose} & \textbf{Exp}   & \textbf{Pose}   & \textbf{Exp}   & \textbf{Pose}  & \textbf{Exp}   & \textbf{Pose} & \textbf{Exp}   & \textbf{Pose} & \textbf{Exp}   & \textbf{Pose} \\
\midrule
 Random & 63.37& 0.10& 67.14&0.10 &2.14 &0.03 &4.06 &3.49 &1.57 &0.01 &0.08 &0.06 \\
Mirror & 69.25& 0.16& 87.11& 0.19& 1.95&0.03 &4.06 &3.54 &1.51 &0.01 &0.74 &0.86 \\
Nearest Motion & 56.00&0.07 & 58.94&0.08 &2.04 &0.02 &3.94 &3.46 &1.62&0.01  & 0.13& 0.13\\
 ELP$*$~\cite{song2023emotional}&22.06 & \textbf{0.01}&22.09 & \textbf{0.01}&0.45& \textbf{0.01}& 5.07&3.64& 0.49&0.01 & 0.09& 0.04 \\
 RLHG~\cite{zhou2022responsive} & 30.50&0.02 &31.74 &0.02  &0.69 &\textbf{0.01} &3.84 &3.65 &0.94 &0.01 &0.09 &0.03     \\
 L2L~\cite{ng2022learning} & 23.56&0.02 &23.94 &0.02 &0.75 &\textbf{0.01} &3.50 &3.00 &0.58 &0.01 &0.12 &0.04\\
L2L$\dagger$~\cite{ng2022learning} &21.63 &0.02 &22.03 &0.02 &0.73 &\textbf{0.01} &3.50 &3.00 &0.67 &0.01 &0.14 &0.03 \\
GT & - & -& - & -& - & - &4.97 & 3.69&0.78 &0.01 &- & - \\
\midrule

\ourmodel$\dagger$  &\textbf{12.49 }&\textbf{0.01} &\textbf{12.95} & 0.02&\textbf{0.29} &\textbf{0.01} &4.53 &2.93 &0.94 & 0.01&\textbf{0.04} &\textbf{0.02}\\
\bottomrule
\end{tabular}}}

\end{table}

\subsection{Benchmark Performances}

\noindent \textbf{Quantitative Results.} We retrain RLHG~\cite{zhou2022responsive}, L2L~\cite{ng2022learning} and ELP~\cite{song2023emotional} on the ViCo dataset~\cite{zhou2022responsive}. Since the \pretraining of our pipeline is conducted on CANDOR~\cite{reece2023candor}, we also pre-train L2L~\cite{ng2022learning}, a recent state-of-the-art method, which adopts a similar approach using VQ-VAE~\cite{van2017neural} in the model, on CANDOR~\cite{reece2023candor}, for fair comparison. Table~\ref{tab:quant_comp_vico} shows the quantitative comparison of Listener Generation on ViCo~\cite{zhou2022responsive}. Based on the metrics presented in Table~\ref{tab:quant_comp_vico}, it is clear that our proposed \ourmodel outperforms existing methods by a significant margin. Our method is about $1.5$ times better than the current state-of-the-art method L2L~\cite{ng2022learning} in terms of the most important metrics \textit{FD} and \textit{P-FD} of expressions in the Listener Generation task. The performance gain of L2L after being trained on CANDOR~\cite{reece2023candor} remains limited compared to the gain of our method, which shows that the effectiveness of \pretraining is as a result of our novel framework rather than additional data. 
In measuring motion diversity (Var, SID), \ourmodel generates diverse listener reactions to the speaker. 
We expect that listeners' movements will showcase a greater range of diversity when the \textit{FD} and \textit{MSE} are kept to a minimum. As demonstrated in Table~\ref{tab:quant_comp_vico}, it is evident that even when FD and MSE are performing well (close to ground truth), \ourmodel remains highly effective in both the Var and SID, which benefits from our proposed \pretraining between speakers and listeners. The RPCC measures the motion synchronization between the speaker and the listener, and it can be found that our method achieves competitive performance.

We also retrain RLHG~\cite{zhou2022responsive}, L2L~\cite{ng2022learning} and our model on LM\_Listener~\cite{ng2023text2listen} dataset. Table.~\ref{tab:quant_comp_lm} demonstrates the quantitative comparison. We can observe that \textit{FD} and \textit{P-FD} of expressions of \ourmodel is about two times better than L2L~\cite{ng2022learning} and  three times better than RLHG~\cite{zhou2022responsive}. We also achieve state-of-the-art in all metrics evaluating the data distribution distances or accuracy between generation and ground-truth while maintaining the ability to generate more diverse motions.
\begin{figure}[t]
    \centering
    \includegraphics[width=\textwidth]{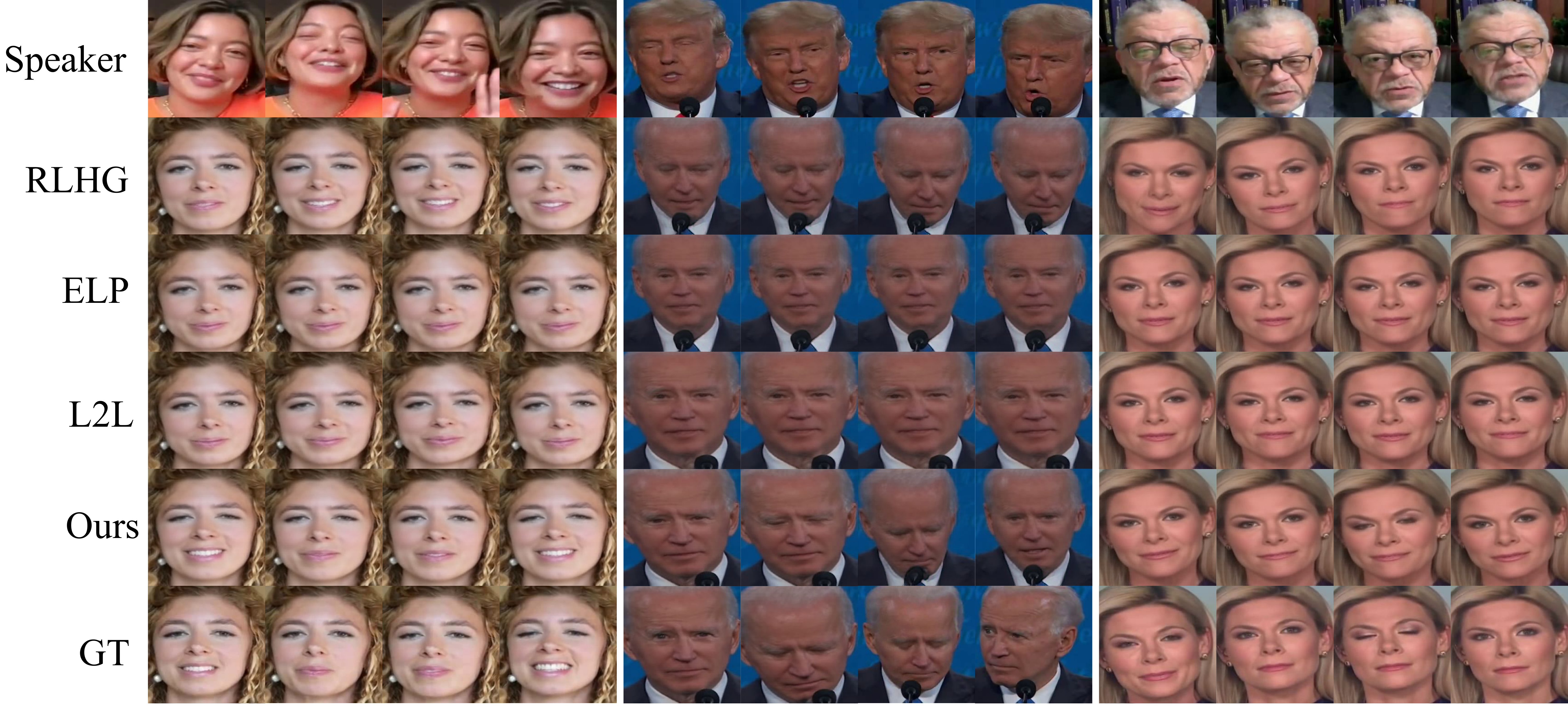}
    \caption{Comparison with L2L~\cite{ng2022learning}(pre-trained on CANDOR~\cite{reece2023candor}), ELP~\cite{song2023emotional}, and RLHG~\cite{ECCV2022}. Our method can generate diverse head movements while maintaining facial expressions that better align with speakers' sentiments.}
    \label{fig:comparison}
\end{figure}
\begin{figure}[t]
    \centering
    \includegraphics[width=\textwidth]{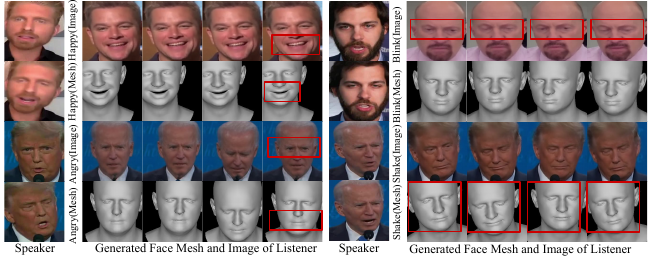}
    \caption{\ourmodel can generate diverse listener emotions (e.g. Happy and Angry) and facial behaviors (e.g. Eyes Blinking and Head Shaking).}

    \label{fig:ours}
\end{figure}
\begin{wraptable}{}{0.5\textwidth}
\centering
\caption{Ablation Analysis of Listener Generation on ViCo~\cite{zhou2022responsive} and LM\_Listener~\cite{ng2023text2listen} Datasets.  VQ represents the VQ-VAEs for speakers and listeners, \pretrainingshort represents \pretraining on CANDOR \cite{reece2023candor}, $Dec^{(l)}_{VQ}$ represents unfreezing listener's VQ decoder, $\mathcal{L}_c$ represents contrastive loss, and S-L represents the two-branch architecture that jointly models speaker and listener behaviors.}
\label{tab:ablation}
\resizebox{0.5\textwidth}{!}
{
{\begin{tabular}{ccccc|cccc}

\toprule
 \multirow{2}[2]{*}{VQ}       &  \multirow{2}[2]{*}{DIM} &  \multirow{2}[2]{*}{$Dec^{(l)}_{VQ}$} &  \multirow{2}[2]{*}{$\mathcal{L}_c$} &  \multirow{2}[2]{*}{S-L} & \multicolumn{2}{c}{ViCo}                        & \multicolumn{2}{c}{LM\_Listener}                \\ 
 &&&&&
\multicolumn{1}{c}{\textbf{MSE}$\downarrow$} & \multicolumn{1}{c}{\textbf{FD}$\downarrow$} & \multicolumn{1}{c}{\textbf{MSE}$\downarrow$} & \multicolumn{1}{c}{\textbf{FD}$\downarrow$} \\
\midrule
\checkmark        & \checkmark        & \checkmark              & \checkmark            & \checkmark          & \multicolumn{1}{c}{\textbf{0.62}}   & \multicolumn{1}{c}{\textbf{24.15}}   & \multicolumn{1}{c}{\textbf{0.27}}   & \multicolumn{1}{c}{\textbf{12.94}}   \\
 \xmark        & \checkmark        & \xmark               & \checkmark            & \checkmark          & 0.92                  & 42.52                & 0.66                  & 33.07                  \\
\checkmark        &    \xmark      & \checkmark           & \checkmark            & \checkmark          & 0.75                  & 33.84                  & 0.60                 & 27.43                 \\
\checkmark        & \checkmark        &            \xmark    & \checkmark            & \checkmark          & 0.81                  & 39.63                  & 0.69                  & 31.93                  \\
\checkmark        & \checkmark        & \checkmark              &      \xmark        & \checkmark          & 0.65                  & 28.47                  & 0.32                  & 17.92                  \\
\checkmark        & \checkmark        & \checkmark              & \xmark            &       \xmark     & 0.72                  & 30.92                 & 0.47                  & 21.27 \\ \bottomrule

\end{tabular}}}

\end{wraptable}
\noindent \textbf{Qualitative Results.} We first compare with the photorealistic results of RLHG \cite{zhou2022responsive} and L2L~\cite{ng2022learning} in Figure~\ref{fig:comparison}. We feed the EMOCA~\cite{EMOCA:CVPR:2021} 3DMM predictions and the listener appearance into PIRenderer~\cite{ren2021pirenderer} and render the corresponding images frame by frame. Since there is no guideline or source code on the renderer's settings in L2L~\cite{ng2022learning}, we use PIRenderer~\cite{wang2018video} official code for training and generation. We fine-tune a unique renderer for each speaker-listener pair and render the listener image sequence using the same fine-tuned weight of each pair for all approaches. Compared to RLHG~\cite{zhou2022responsive}, L2L~\cite{ng2022learning}, and ELP~\cite{song2023emotional}: 
1) our method demonstrates more expressive behaviors that synchronize with speakers' and target listeners' sentiments (first example), 2) our method can generate diverse head movements compared to other methods (second example).
We also demonstrate our method's ability to model emotions and facial behaviors of our model in Figure~\ref{fig:ours}. 
In particular, we show that \ourmodel can generate multiple types of social signals such as smiles, head shaking and blinks. These suggest the benefits of \pretraining of \ourmodel on the large-scale CANDOR~\cite{reece2023candor} dataset with more diverse dyadic behaviors. 

\subsection{Ablation Analysis}

Table~\ref{tab:ablation} illustrates the contribution of each component to \ourmodel. Initially, we note that VQ-VAE stands out as the most crucial module in our framework. Furthermore, our observation that directly predicting listener's motions as continuous signals yields less diverse motions aligns with prior research in L2L \cite{ng2022learning} and ELP \cite{song2023emotional}. As expected, \pretraining on large-scale dyadic behaviors significantly enhances performance for both ViCo and LM\_Listener datasets. In contrast to the pre-training approach in L2L \cite{ng2022learning}

 that fixes the VQ-Decoder to decode continuous signals from discrete motion predictions, integrating the VQ-Decoder into the listener motion prediction model notably enhances performance, ranking as the second most important component in \ourmodel. While $\mathcal{L}_c$ ranks as the least important component, it still contributes significantly to performance by encouraging role-specific encoders to extract meaningful information from input signals relevant not only to the current person but also the conversational partner. Finally, we demonstrate the benefits of jointly modeling both speaker and listener behaviors with $S-L$, which aligns with existing literature indicating that speaker and listener behaviors correlate and mutually influence each other. we also provide modality ablation as well as online listener behavior generation results in the supplementary material.

\subsection{Speaker Generation}

\begin{figure}[t]
  \begin{center}
    \includegraphics[width=0.7\textwidth]{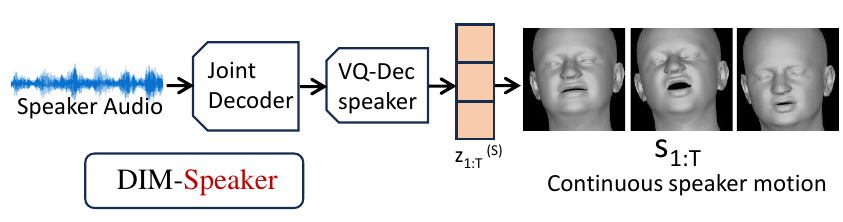}
  \end{center}
  \caption{For Speech-Driven Speaker Generation, we fine-tune the pre-trained weights of the joint decoder and speaker VQ-Decoder from \pretrainingshort. The generated speaker's motion from \textsc{DIM-Speaker} is optimized by the same cross-entropy and reconstruction loss as \ourmodel. }
  \label{fig:speaker}
\end{figure}
\begin{wraptable}{}{0.5\textwidth}
\centering
\caption{Quantitative Comparison of Speech-Driven Speaker Generation on BiWi~\cite{fanelli2010} Dataset.}
\label{tab:quant_comp_biwi}
\scalebox{0.9}{
\begin{tabular}{lcc}
\toprule
\multirow{2}[2]{*}{Method} & \textbf{LVE$\downarrow$} & \textbf{FDD$\downarrow$} \\
&$(\times 10^{-4})$ &$(\times 10^{-4})$\\
\midrule
  MeshTalk~\cite{richard2021meshtalk} &6.14 & 2.53  \\
  FaceFormer~\cite{fan2022faceformer}  &5.40 &1.37  \\
  CodeTalker~\cite{xing2023codetalker} & 5.93 &2.42  \\

\textsc{DIM-Speaker}  & \textbf{3.86} & \textbf{0.98} \\
\bottomrule
\end{tabular}}
\end{wraptable} 
As mentioned in the previous sections, since we introduce a dyadic framework capturing contextualized interactions among listeners and speakers, our approach can also animate speaker motion from the speaker's speech input. We name it \textsc{DIM-Speaker}, and the fine-tuning pipeline is shown in Figure~\ref{fig:speaker}. For this task, we compared the results to several state-of-the-arts methods, e.g. MeshTalk~\cite{richard2021meshtalk}, FaceFormer~\cite{fan2022faceformer} and CodeTalker~\cite{xing2023codetalker} on BiWi~\cite{fanelli2010} dataset.

We retrain these methods and compare the output motion parameters in the FLAME point cloud space, which consists of 5023 vertices converted from EMOCA 3DMMs. We follow the past work and adopt \textit{Lip Vertex Error (LVE)}, which measures the discrepancy in lip movement for a generated sequence compared to the ground truth by calculating the maximum Euclidean (L2) distance error for the lip vertices in each frame, averaged over all frames, and \textit{upper-face dynamics deviation(FDD)}, which evaluates the variation of facial dynamics for a motion sequence, as evaluation metrics. 

As shown in Table.~\ref{tab:quant_comp_biwi}, we outperform previous methods in this evaluation space, which demonstrated the effectiveness of our framework for the Speaker Generation task. 


%% file: sec/5_conclusion.tex
\section{Conclusion}\label{sec:conclusion}


In this study, we introduced \pretraining, a novel self-supervised pre-training strategy designed to improve the model's capability to effectly encode representations from both speaker and listener motions via large-scale dyadic data. Our comprehensive experiments and detailed visualizations demonstrate the capability of \textsc{DIM-Listener} and \textsc{DIM-Speaker} in generating accurate and realistic motions for both listeners and speakers.


%% file: X_suppl.tex
\title{Supplementary materials for DIM: Dyadic Interaction Modeling for Social Behavior Generation} 


\author{Minh Tran$^*$ \orcidlink{0009-0004-2391-3563} \and
Di Chang$^*$ \orcidlink{0009-0002-0281-8896} \and
Maksim Siniukov\and
Mohammad Soleymani\orcidlink{0000-0002-5873-1434}}

\authorrunning{M.~Tran et al.}

\institute{University of Southern California, Institute for Creative Technologies \\ \email{soleymani@ict.usc.edu}\\}

\maketitle
\def\thefootnote{*}\footnotetext{equal contribution}
\def\modelname{Ex2Eg-MAE }
\def\modulename{Perspective Shift Estimation Module }

\appendix


\section{Metrics for evaluation}
In this section, we further define our metrics in detail as a supplement to the main text. 
\begin{itemize}
    \item \textit{Frechet Distance (FD)}:  Motion realism is evaluated through the distributional discrepancy between the generated and actual ground-truth motions. This is quantified by calculating the Fréchet Distance (FD)~\cite{heusel2017gans} within the domains of facial expressions and head poses across the entire motion sequence.

  \item\textit{Paired FD for synchrony (P-FD)}: Listener-speaker dynamics quality is based on the distances between distributions of listener-speaker pairs. We concatenate the generated listener motions with speaker motions and calculate Fréchet Distance (FD)~\cite{heusel2017gans}.  

  \item\textit{Mean Squared Error (MSE)}: MSE between generated and ground-truth motions.

  \item\textit{SI for Diversity (SID)}: In L2L~\cite{ng2022learning}, the diversity of predictions of facial and head motions is assessed using k-means clustering. This approach facilitates the evaluation of variation within predicted results by calculating the entropy of the histogram of cluster ids within each sequence.

  \item\textit{Variance (Var)}: Following L2L~\cite{ng2022learning}, this metric measures the variance of the listener motions on time series. 

  \item\textit{Residual Pearson Correlation Coefficient (rPCC)}: The Pearson Correlation Coefficient (PCC)~\cite{messinger2009automated, riehle2017quantifying}, evaluates the motion of the face and head on a frame-by-frame basis, indicating how the listener's movements correlate with those of the speaking individuals. The $L_1$ distance is utilized to quantify the discrepancy between the generated PCC and the ground truth PCC.
\end{itemize}

\section{Modality Ablation \& Online Behavior Generation}
We present the results of our modality ablation and online behavior generation experiments for DIM-Listener on ViCo in Table \ref{tab:ablation_supp}. In the modality ablation experiments, we mask all inputs of a modality to zeros. For the online generation experiments, we add causal masks during the fine-tuning phase and reuse the pre-trained weights. The online generation results (which are slightly worse but still superior to prior work) are also reported in Table \ref{tab:ablation_supp} for ViCo.
\begin{table}[t]
\centering
\caption{Modality contribution \& online gen. experiment on ViCo.}
\scalebox{0.8}
{\begin{tabular}{lcccccc||cc}
    & \multicolumn{2}{c}{V} & \multicolumn{2}{c}{A} & \multicolumn{2}{c}{A+V} & \multicolumn{2}{c}{Online Gen.} \\ 
    & Exp       & Pose      & Exp       & Pose      & Exp        & Pose & Exp        & Pose       \\ \hline
MSE & 0.85         & 0.01         &  0.97        & 0.02         & 0.7        & 0.01 & 0.79        & 0.01       \\
FD  & 32.8         & 0.07         & 46.9         & 0.08         & 23.9       & 0.06  & 28.4        & 0.06  \\ \hline  
\end{tabular}
}
\vspace{-15pt}
\label{tab:ablation_supp}
\end{table}
\section{Generated Listener Motions User Study}
We asked 50 participants from \textbf{Prolific}, a crowdsourcing platform, to rate the quality of \textbf{eight} generated listeners in the test set of ViCo~\cite{zhou2022responsive} from L2L~\cite{ng2022learning}, RLHG~\cite{zhou2022responsive}, ELP~\cite{song2023emotional}, and our method on a scale of 0(low) to 5(high). The methods were anonymized and randomly ordered for each question. The result is available in Table~\ref{tab:user_study_rebuttal}. The average duration of the videos is $14$ seconds. Criteria for judgment: 1) The response from the listener provides correct feedback to the speaker. 2) The head motions and facial expressions are diverse and smooth.\\
\begin{table}[h]
\centering
\caption{User Study Results. Scale: 0-5; the higher, the better.
}
\scalebox{0.85}
{\begin{tabular}{lcccccccccccc}
\toprule
Method  & {\textbf{Video-1}} & {\textbf{Video-2}}& {\textbf{Video-3}}& {\textbf{Video-4}}&{\textbf{Video-5}} & {\textbf{Video-6}} & {\textbf{Video-7}}& {\textbf{Video-8}}& {\textbf{Average}}\\
\midrule
ELP~[62]  &2.12 &2.16 & 1.98&2.02 &2.00 &1.88 & 1.80& 1.88&1.98 \\
RLHG~[77]  & 2.45& 2.67& 2.80&2.75 & 2.74& 2.18& 2.43& 2.12& 2.51\\
 L2L~[47]  & 2.92&2.78 &2.76 &3.20 & 3.69&2.14 &2.96 & 2.63 & 2.89\\
Ours  & \textbf{3.90} & \textbf{3.72}& \textbf{3.98}&\textbf{3.47} &\textbf{3.98} &\textbf{ 2.94} & \textbf{3.75}& \textbf{3.31} & \textbf{3.64}\\
\bottomrule
\end{tabular}}
\label{tab:user_study_rebuttal}

\end{table}

\section{Limitations and Future Works}
\subsection{Rendering Network}
In our framework, we observed that the image quality produced by PIRenderer~\cite{ren2021pirenderer} does not always meet our expectations. Specifically, the rendered images exhibit distortions for certain individuals. Moreover, we identified a crucial need for individualized fine-tuning of the rendering network for each subject. This necessity stems from the model's inability to effectively generalize across different identities, even after training. Consequently, we have adopted the strategy of training a distinct PIRenderer~\cite{ren2021pirenderer} model for every individual in our test set. Future research could explore the integration of a more adaptable rendering model into the framework, to enhance the photorealism of the generated video.

\subsection{EMOCA/FLAME Representation}

As mentioned in the paper,  we adopt EMOCA~\cite{EMOCA:CVPR:2021} to represent facial expressions and head pose motions, encapsulated in a compact output dimensionality of $[56,]$, specifically $[50,]$ for facial expressions and $[6,]$ for head poses. In the context of our speaker generation task, we demonstrate the flexibility of our model by converting this prediction into a FLAME~\cite{FLAME:SiggraphAsia2017} point cloud representation, comprising 5023 vertices. This approach contrasts with prior models that employ a more detailed 3D face geometry with 23370 vertices~\cite{fan2022faceformer,richard2021meshtalk,xing2023codetalker}. While EMOCA 3DMMs offer a streamlined representation, they inherently sacrifice some details in capturing facial expressions and head movements. To enhance the fidelity of listener and speaker generation tasks in future research, we suggest exploring more nuanced and expressive representations that more accurately reflect facial behaviors.